# Novel techniques for improving NNetEn entropy calculation for short and noisy time series

**Authors:** Hanif Heidari[1], Andrei Velichko[2], Murugappan Murugappan[3,4,5*] and Muhammad E.H. Chowdhury[6]

**Affiliations:**

1. Hanif Heidari, Department of Applied Mathematics, Damghan University, Damghan, Iran

2. Andrei Velichko, Institute of Physics and Technology, Petrozavodsk State University, 31 Lenina Str., 185910 Petrozavodsk, Russia

3. Murugappan Murugappan, Department of Electronics and Communication Engineering, Kuwait College of Science and Technology, Block 4, Doha, 13133, Kuwait

4. Murugappan Murugappan, Department of Electronics and Communication Engineering, Vels Institute of Sciences, Technology, and Advanced Studies, Chennai, India

5. Center for Unmanned Aerial Vehicles (CoUAV), Universiti Malaysia Perlis, Arau, 02600, Perlis, Malaysia

6. Muhammad E.H.Chowdhury, Department of Electrical Engineering, College of Engineering, Qatar University, Doha 2713, Qatar

**Corresponding Author:**

Murugappan Murugappan,

Full Professor,

Department of Electronics and Communication Engineering,

Kuwait College of Science and Technology,

Block 4, Doha, 13133,

Kuwait

Email ID: m.murugappan@kcst.edu.kw, m.murugappan@gmail.com

Phone: +965-24972853

# Novel techniques for improving NNetEn entropy calculation for short and noisy time series

**Abstract**

Entropy is a fundamental concept in the field of information theory. During measurement, conventional entropy measures are susceptible to length and amplitude changes in time series. A new entropy metric, neural network entropy (NNetEn), has been developed to overcome these limitations. NNetEn entropy is computed using a modified LogNNet neural network classification model. The algorithm contains a reservoir matrix of N = 19625 elements that must be filled with the given data. A substantial number of practical time series have fewer elements than 19625. The contribution of this paper is threefold. Firstly, this work investigates different methods of filling the reservoir with time series (signal) elements. The reservoir filling method determines the accuracy of the entropy estimation by convolution of the study time series and LogNNet test data. The present study proposes 6 methods for filling the reservoir for time series of any length 5 ≤ N ≤ 19625. Two of them (Method 3 and Method 6) employ the novel approach of stretching the time series to create intermediate elements that complement it, but do not change its dynamics. The most reliable methods for short time series are Method 3 and Method 5. The second part of the study examines the influence of noise and constant bias on entropy values. In addition to external noise, the hyperparameter (bias) used in entropy calculation also plays a critical role. Our study examines three different time series data types (chaotic, periodic, and binary) with different dynamic properties, Signal to Noise Ratio (SNR), and offsets. The NNetEn entropy calculation errors are less than 10% when SNR is greater than 30 dB, and entropy decreases with an increase in the bias component. The third part of the article analyzes real-time biosignal EEG data collected from emotion recognition experiments. The NNetEn measures show robustness under low-amplitude noise using various filters. Thus, NNetEn measures entropy effectively when applied to real-world environments with ambient noise, white noise, and 1/f noise.

## 1. Introduction

Entropy-based methods are efficient tools for analyzing any nonlinear, non-stationary, or dynamic signals [1, 2]. Compared to other nonlinear methods [3], entropy-based feature analysis is computationally efficient and highly suitable for analyzing the input data with fewer to a larger number of samples. The entropy-based analysis has been widely used in medical diagnosis, fault detection, image processing applications, information encryption and speech recognition [4–9]. Researchers are actively utilizing entropy measures in many real-world applications due to their efficient information retrieval from the input data. Yağ at al. used entropy-based features to classify plant leaf diseases [10]. Multiscale symbolic fuzzy entropy is used for fault detection in rotating machinery [11]; Minhas et al. examined the weighted entropy method for bearing fault detection [2]; Ai et al. studied the fusion information entropy for roller bearing fault detection in aircraft engines [4]; Ra et al. evaluated the permutation entropy of Electroencephalogram (EEG) signals for early diagnosis of epileptic seizures [5]; Zavala-Yoe et al. investigated the multiscale entropy to study Doose and Lennox-Gastaut syndromes in children [6]; Benedetto et al. utilized the Shannon entropy to model the flows between different financial time series [12]; Silva et al. investigated the predictability of monthly precipitation time series using permutation entropy; Nie et al. introduced a generalized entropy measure for image segmentation [13]; and Oludehinwa et al. investigated the relationship between the degree of complexity of magnetospheric dynamics and various categories of geomagnetic storms using the maximum Lyapunov exponent and the approximate entropy measure [14]. The major challenges in conventional entropy measurements are: (a) hyperparameter tuning; (b) input data length limitations; (c) the effect of input noise on the results.

Recently, Velichko and Heidari proposed a new measure of entropy (NNetEn) based on the LogNNet artificial neural network (ANN) [15]. In the analysis of time series, the number of epochs is the only control parameter that achieves substantially more stable results than other conventional entropy measures (approximate entropy, Shannon entropy, etc). Further, it has the advantage of being independent of signal amplitude, suitable for analyzing short and lengthy-time series data, and does not consider the probability distribution within the data, which makes it distinct from other traditional entropy measures [15]. The main objective of this work is to propose a novel ANN-based entropy calculation for variable data lengths with limited number of hyperparameters to overcome constraints of conventional entropy computations. Hence, this paper investigates different reservoir matrix filling methods and the effects of noise on the value of NNetEn in light of its advantages over other entropy measures.

In the first part of the scientific results, a method is presented for filling the reservoir with time series data (signals). Ultimately, the adequacy of the entropy estimate is determined by the reservoir filling method incorporated into the convolution of the time series under study with LogNNet test data derived from the MNIST database. In this study, six methods (Methods 1-6) are proposed for filling a reservoir of time series with a length of $5 \leq N \leq 19625$. A range of N in this range is most commonly used for financial markets [16], physical experiments [17, 18], and biological and medical data [19, 20]. Both Method 3 and Method 6 use a novel approach to stretching time series in which intermediate elements are generated to complement the time series without changing its dynamics. In particular, this approach can be applied to short series (N < 1000) of physical data. We believe that Method 3 and Method 5 are the most effective methods for short time series. A number of studies have already proven the effectiveness of these methods, as indicated in the preprint of this work. An analysis of ship-radiated noise signals was conducted by Li et al. using NNetEn [21] (Method 5). A method of detecting motor imaginary in patients with stroke or spinal cord injury was developed by Heilari using NNetEn [22]. According to his findings, eight channels are enough for classification, unlike previous studies that considered 30 channels. In remote sensing imagery and geophysical mapping, Velichko et al. used Method 1 for two-dimensional NNetEn. An S-Switch-based chaotic spike oscillator circuit was analyzed with Method 3 for bifurcation and entropy analysis [25]. During major geomagnetic storms, Oludehinwa et al. used Method 3 for NNetEn to examine the response of dynamical complexity in traveling ionospheric disturbances across Eastern Africa sector. Thus, in the first part of the work, we examined different matrix-filling techniques for entropy computation of short time series. The experimental results demonstrate that the proposed matrix filling method can be applied to time series with a length of $N \geq 5$.

The second part of the study is devoted to the problem of the influence on the entropy value of the presence of noise or a constant bias in the signal. Results demonstrate that the error in calculating the NNetEn entropy is less than 10% when the signal-to-noise ratio is greater than 30dB. Also shown, with an increase in the bias component, the role of the chaotic components is weakened, and entropy decreases. In relation to comparing the performance of different entropy measures, noise can be regarded as one of the most significant factors affecting the entropy value [27]. It has been demonstrated that Kolmogorov-Sinai entropy and the correlation integral algorithm are inefficient methods for calculating entropy in the presence of noise [28]. It has been observed that the permutation entropy measure is highly volatile in the presence of noise [29] . Similarly, the entropy features are generally calculated from the given time series following the removal of noise (filtering). Consequently, reducing the effects of noise in time series data has become a popular research topic over the last few decades. In [30], Xie and Guo applied fuzzy spectrum entropy analysis to denoise biomedical signals; Chatterjee et.al. reduced the effect of noise from chemical and electronic sources using patterns of ion current chromatograms [31]; Na et.al. compared cross-correlation with Shannon entropy in order to minimize the effects due to unwanted noise [32]; Wang et.al. studied multi-fault features of transmission in the presence of noise [33]. Time series length is also an impacting factor that affects entropy [27, 34, 35]. Conventional entropy measures are not suitable for time series analysis with a limited number of data. This is because they are very sensitive to the length of the data series [36, 37]. Wu et al. presented a modification to multiscale entropy to deal with imprecise entropy values [38], and Niu and Wang used a simplified version of multiscale entropy to analyze the short financial time series to overcome this challenge [36].

We present a real-time example of time-series data collected from an emotion recognition experiment for testing the efficacy of the proposed NNetEn on time-series data analysis in the third part of the article. Affective computing has become a hot topic from both a theoretical and practical perspective in recent years due to rapid technological advancements. Over the past few years, affective computing research has grown rapidly and has been applied to healthcare, robotics, management, marketing, and smart technology [39, 40]. The manifestation of emotion can be determined by a variety of means, including facial expressions, gestures, speech signals, and biosignals such as electroencephalograms (EEGs), electromyograms (EMGs), electrocardiograms (ECGs), etc. Due to their ability to capture minute changes in emotions, EEG signals are more robust than other signals because they are not subjected to fake or masking behaviors. The acquisition of EEG signals is inexpensive, has a high temporal resolution, and offers an adequate spatial resolution. A study by Soroush et al. [39] examined different methods for classifying emotions based on EEG signals. In their study [43], Murugappan et al. used extreme machine learning combined with recurrent quantification analysis (RQA) to detect emotion in EEG data collected from Parkinson's patients (PD). A mean differential entropy method is presented by Li et al for classifying emotional states in EEG data [44]. A tunable Q-wavelet transform was implemented by Murugappan et al. [42] to improve the accuracy of EEG emotion recognition in PD patients. A tunable Q-wavelet transform, and support vector machine method were used by Tuncer et al. to recognize emotions based on EEG signals [40]. A study by Li et al attempted to identify emotions through EEG in the presence of noise [45]. Researchers collected EEG data from 14 healthy subjects for the detection of six basic emotions, namely happiness, sadness, fear, anger, disgust, and surprise [46]. We examined real-time data from a normal control subject using a 14-channel Emotiv EPOC wireless EEG data acquisition device in order to evaluate NNetEn's performance. A variety of filters are applied to the signal to demonstrate that NNetEn measures are robust under low-amplitude noise. Therefore, NNetEn is an effective measure of entropy in real-world environments that involves ambient noise, white noise, and 1/f noise.

The major contributions in the present work are:
- Proposed six different types of matrix filling (reservoir) methods to compute the NNetEn on the input data with variable data length.
- Investigated the performance of NNetEn under different types of noises and constant bias.
- Analyzed the performance of NNetEn on real-time physiological signals (EEG).

The rest of the paper is structured as follows. In Section 2, the LogNNet model and matrix padding techniques are presented. In Section 3, numerical examples are presented to investigate the efficiency of the matrix filling techniques and the effect of noise on the NNetEn measure. The performance of NNetEn with real-time EEG data is also presented in Section 3. Finally, Section 4 is devoted to conclusions.

## 2. Methods

### 2.1. LogNNet model for entropy calculation

NNetEn is a recently introduced entropy measure based on the LogNNet model (Figure 1) [15]. Unlike conventional entropy measures, NNetEn does not consider a probability distribution. In addition, it depends on only one parameter i.e. Number of epochs in the LogNNet model. These advantageous features make NNetEn a powerful entropy measure for any time series data analysis. In the NNetEn algorithm, the Modified National Institute of Standards and Technology (MNIST-10 [47]) dataset is considered the input of LogNNet. MNIST-10 is a popular benchmark and large size dataset which is widely used for comparing the efficiency of different machine learning algorithms [48].This dataset is divided into training and testing parts which contain 60000 and 100000 images, respectively with a resolution of 28 × 28 pixels. The NNetEn algorithm requires a reservoir matrix which is constructed by using the given time series. The reservoir matrix is denoted by $W_1$ in Figure 1, where P is the number of considered neurons. In the classification phase, a feedforward artificial neural network with single layer and 25 neurons is used. Therefore, the reservoir matrix has a capacity to accommodate $N_0 = 19625$ elements.

.

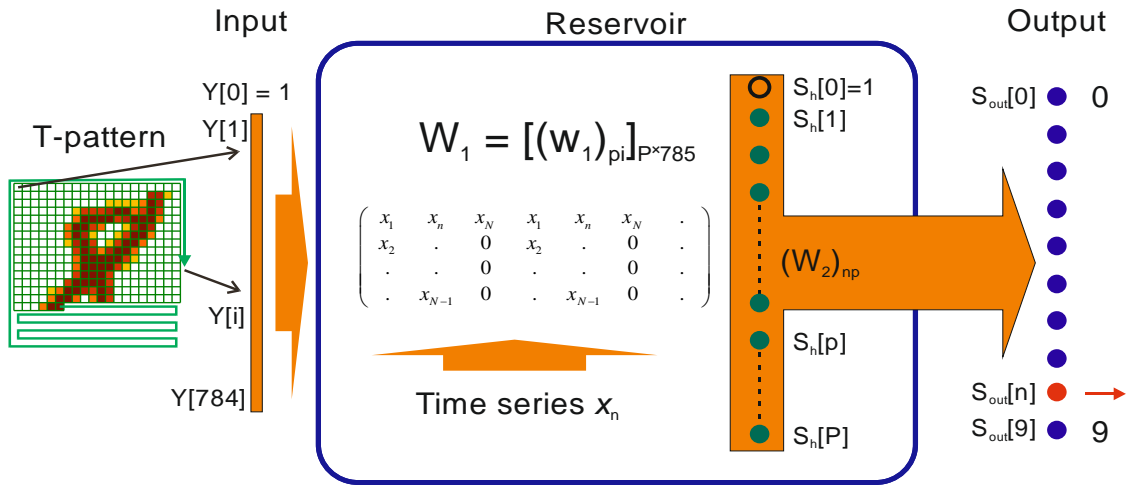

**Figure 1.** The LogNNet model structure for NNetEn calculation [49]

The number of epochs $Ep$ is the only control parameter in the LogNNet algorithm. The classification accuracy of LogNNet shows the degree of chaos in the reservoir matrix and is defined as the NNetEn entropy measure

$$\text{NNetEn}(Ep) = \frac{Classification\ accuracy\ \%}{100} \tag{1}$$

The flowchart of the NNetEn algorithm is shown in Table 1.

**Table 1.** NNetEn algorithm [15].

| |
|---|
| 1. Loading time series $x_n$. |
| 2. Loading the MNIST-10 database and the T-pattern-3 pattern [50] to convert input images into the vectors array $Y$. |
| 3. Initializing the initial values of weights and neurons. |
| 4. Constructing the reservoir matrix $W_1$ using the given time series (Section 2.3.). |
| 5. Calculating the coefficients for normalization. |
| 6. Determining the number of training epochs $Ep$. |
| 7. Performing the training process of the LogNNet 784:25:10 network using a training set. |
| 8. Performing the testing process of the LogNNet 784:25:10 network with a test set and calculating the classification accuracy using Eq. (1). |

In this article, based on the heuristic approach, we considered the maximum value of the epoch, $Ep = 100$ and NNetEn is referred as NNetEn (100).

The concept of learning inertia (LI) is introduced in [15] for investigating the effect of $Ep$ on the NNetEn value. For a given time series, the learning inertia concerning epochs numbers $Ep1$ and $Ep2$ is defined as follows,

$$LI(Ep1/Ep2) = \frac{\text{NNetEn}(Ep2 \text{ epoch}) - \text{NNetEn}(Ep1 \text{ epoch})}{\text{NNetEn}(Ep2 \text{ epoch})}. \qquad (2)$$

An analysis of the behavior of LI provides valuable information about the convergence of the learning process in neural networks models, which helps the researchers to discover additional features of chaotic signals. For example in [26], LI was used for filtering the NNetEn signal ($Ep1 = 20$, $Ep2 = 1$), and in [15], LI was the feature of the occurrence of weak chaos in the periodic component ($Ep1 = 100$, $Ep2 = 400$). Hence, based on the heuristic approach, we have used the value of Ep1 and Ep2 as 100 and 400, respectively on this work. The calculation of one time series with 100 epochs takes about 37 s using one Intel™ Co™ m3-8100Y CPU @ 1.10 GHz processor thread. The NNetEn calculation program is available for free download (see Section Data Availability).

## 2.3 Reservoir Filling Techniques

In practice, it is difficult to obtain time series with $N = 19625$ data points. Presume that we have a time series data with $N$ data points. If $N > 19625$, we must ignore $N$-19625 data points of the time series and fill the reservoir matrix with the remaining data points. This case rarely occurs in real-world applications. If the total number of data points in the time series data is less than 19625 ($N$) data points, we propose six different types of reservoir filling methods to utilize the LogNNet to compute the NNetEn values. Here, the process of filling is performed through either row-wise or column wise filling of the reservoir matrix. For the purpose of describing the filling processes, we considered a smaller size matrix $W_1$ filled with series $x_n = n$, where $n$ varies from 1 to 9.

   (a) Method 1: Row-wise filling with duplication

In this method, the matrix $W_1$ is filled row by row, doubling the row for subsequent data points, as described in Figure 2.

$$\begin{pmatrix} x_1 & x_2 & . & . & x_n & . & . \\ x_{N-1} & x_N & x_1 & x_2 & . & . & x_n \\ . & . & x_{N-1} & x_N & x_1 & x_2 & . \\ . & . & x_n & . & . & x_{N-1} & x_N \end{pmatrix} \qquad \begin{pmatrix} 1 \to 2 \to 3 \to 4 \to 5 \to 6 \to 7 \\ 8 \to 9 \; 1 \; 2 \; 3 \; 4 \; 5 \\ 6 \; 7 \; 8 \; 9 \; 1 \; 2 \; 3 \\ 4 \; 5 \; 6 \; 7 \; 8 \; 9 \; 1 \end{pmatrix}$$

(a)                  (b)

**Figure 2.** (a) The structure of matrix filling method 1; (b) An example of a time series with nine data points.

(b) Method 2: Row-wise filling with an additional zero element

This method is similar to filling the matrix row by row. We considered the zero values for the remaining data points in the matrix row if the row cannot be filled with all the given time series data points. A schematic description and an example of this method is given in Figure 3.

$$\begin{pmatrix} x_1 & x_2 & . & . & x_n & . & . \\ x_{N-1} & x_N & 0 & 0 & 0 & 0 & 0 \\ x_1 & x_2 & . & . & x_n & . & . \\ x_{N-1} & x_N & 0 & 0 & 0 & 0 & 0 \end{pmatrix} \qquad \begin{pmatrix} 1 \to 2 \to 3 \to 4 \to 5 \to 6 \to 7 \\ 8 \to 9 \; 0 \; 0 \; 0 \; 0 \; 0 \\ 1 \; 2 \; 3 \; 4 \; 5 \; 6 \; 7 \\ 8 \; 9 \; 0 \; 0 \; 0 \; 0 \; 0 \end{pmatrix}$$

(a)                  (b)

**Figure 3.** (a) The structure of matrix filling method 2; (b) An example of a time series with nine data points.

(c) Method 3: Row-wise filling with time series stretching

This method fills the matrix row-wise and stretches the time series to $N = 19625$. The stretching method constructs a new time series $\{z_n\}$ with $N = 19625$ data points, whose data points are linear approximations between two adjacent points of the original time series $\{x_n\}$. Figure 4(a) shows a time series $\{x_n\}$ with $N = 100$ values. The stretched time series $\{z_n\}$ is not unique. It gives almost the same results if we consider the time series as a record of a dynamic process (Figure 4(b)). The detailed arrangement of the data points $\{z_n\}$ near $x_{28}$ is shown in Figure 4(c).

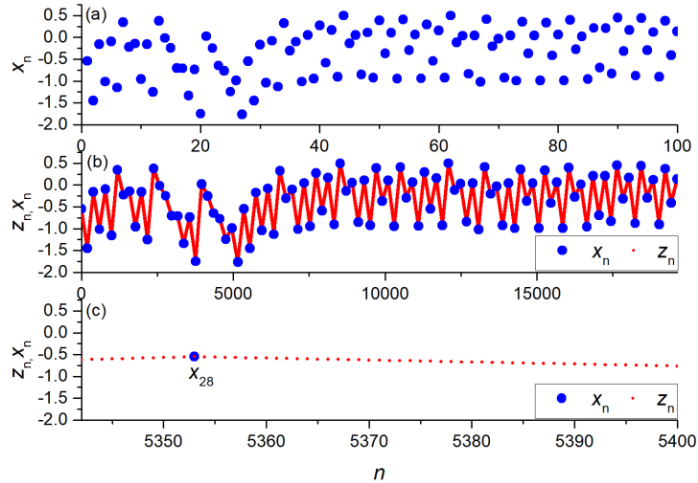

**Figure 4.** (a) The time series $\{x_n\}$ with $N = 100$ data points; (b) Stretching $\{x_n\}$ into the time series $\{z_n\}$ with $N = 19625$ data points; (c) The detailed arrangement of data points $\{z_n\}$ near $x_{28}$.

(d) Method 4: Column-wise filling with duplication

In this method, the matrix $W_1$ is filled column by column, duplicating the columns for the subsequent data points, as shown in Figure 5.

$$\begin{pmatrix} x_1 & x_n & x_N & \cdot & \cdot & \cdot & \cdot \\ x_2 & \cdot & x_1 & x_{N-1} & \cdot & \cdot & \cdot \\ \cdot & \cdot & x_2 & x_N & \cdot & \cdot & \cdot \\ \cdot & x_{N-1} & \cdot & \cdot & \cdot & \cdot & \cdot \end{pmatrix} \quad \begin{pmatrix} 1 & 5 & 9 & 4 & 8 & 3 & 7 \\ 2 & 6 & 1 & 5 & 9 & 4 & 8 \\ 3 & 7 & 2 & 6 & 1 & 5 & 9 \\ 4 & 8 & 3 & 7 & 2 & 6 & 1 \end{pmatrix}$$

(a)                      (b)

**Figure 5.** (a) The structure of matrix filling method 4; (b) An example of a time series with nine data points.

(e) Method 5: Column-wise filling with an additional zero element

This method is similar to Method 2 where the column is considered instead of rows (Figure 6).

$$\begin{pmatrix} x_1 & x_n & x_N & x_1 & x_n & x_N & \cdot \\ x_2 & \cdot & 0 & x_2 & \cdot & 0 & \cdot \\ \cdot & \cdot & 0 & \cdot & \cdot & 0 & \cdot \\ \cdot & x_{N-1} & 0 & \cdot & x_{N-1} & 0 & \cdot \end{pmatrix} \quad \begin{pmatrix} 1 & 5 & 9 & 1 & 5 & 9 & 1 \\ 2 & 6 & 0 & 2 & 6 & 0 & 2 \\ 3 & 7 & 0 & 3 & 7 & 0 & 3 \\ 4 & 8 & 0 & 4 & 8 & 0 & 4 \end{pmatrix}$$

(a)                      (b)

**Figure 6.** (a) The structure of matrix filling method 5; (b) An example of a time series with nine data points.

(f) Method 6: Column-wise filling with time series stretching

In this method, the matrix is filled column by column, stretching the time series to $N = 19625$, as described in method 3.

## 3. Experimental results and discussion

### 3.1. The effect of the length of the time series on the value of NNetEn

We investigated the efficiency and stability of the six matrix-filling methods using the above-mentioned matrix-filling methods on three types of time series, namely periodic discrete maps, binary discrete maps, and logistic maps, taking into consideration the total number of data points $N$. A robust analysis of different matric-filling methods can be obtained by considering periodic [51], binary [52], and chaotic time series [51] as follows:

(i)     Periodic discrete map:

$$x_n = A \cdot \sin\left(\frac{n \cdot 20\pi}{19625}\right) \tag{3}$$

(ii)    Binary discrete map:

$$x_n = n \bmod 2,$$

$$x_n = (1,0,1,0,1,0,1,0,1,\ldots\ldots) \tag{4}$$

(iii) Logistic map:

$$x_{n+1} = r \cdot x_n \cdot (1 - x_n) \tag{5}$$

The epochs number ($n$) is set to 100 in this sub-section.

Figure 7 shows different matrix filling methods applied to measure the NNetEn values for a chaotic logistic map with $r = 3.8$ [51] and by varying the number of data points $N$ (i.e $N = 10,\ldots..,19625$). The dashed red line shows the reference level corresponding to NNetEn at $N = 19625$. If N is reduced below 19625, there should not be a significant change in entropy (NNetEn) compared to the reference level.

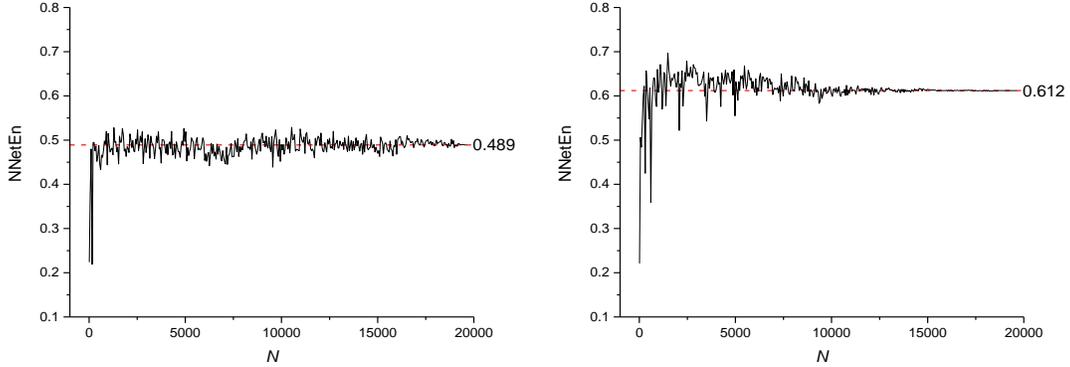

(a) (d)

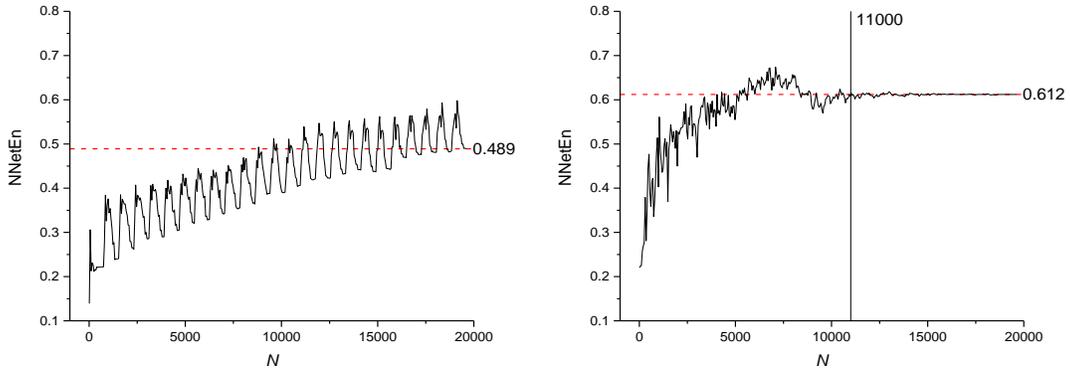

(b) (e)

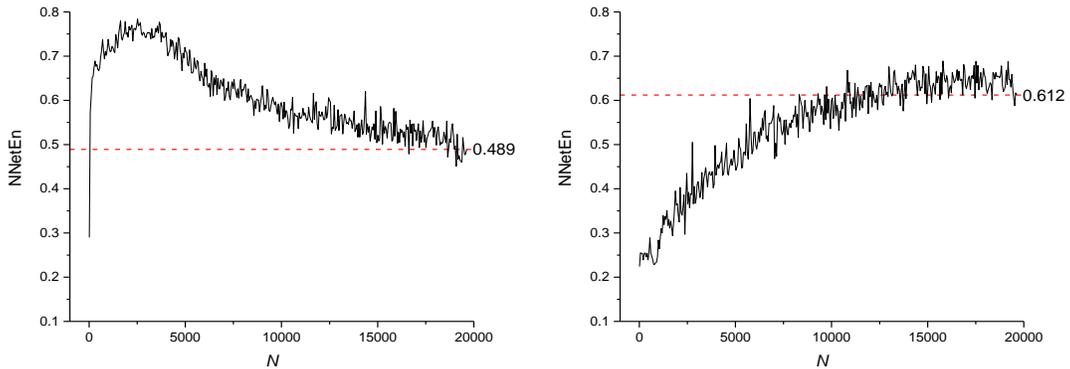

(c) (f)

**Figure 7.** The effect of time series length on NNetEn values for the logistic map (Eqn 5) using different matrix filling methods: (a) Method 1 (b) Method 2 (c) Method 3 (d) Method 4 (e) Method 5 (f) Method 6.

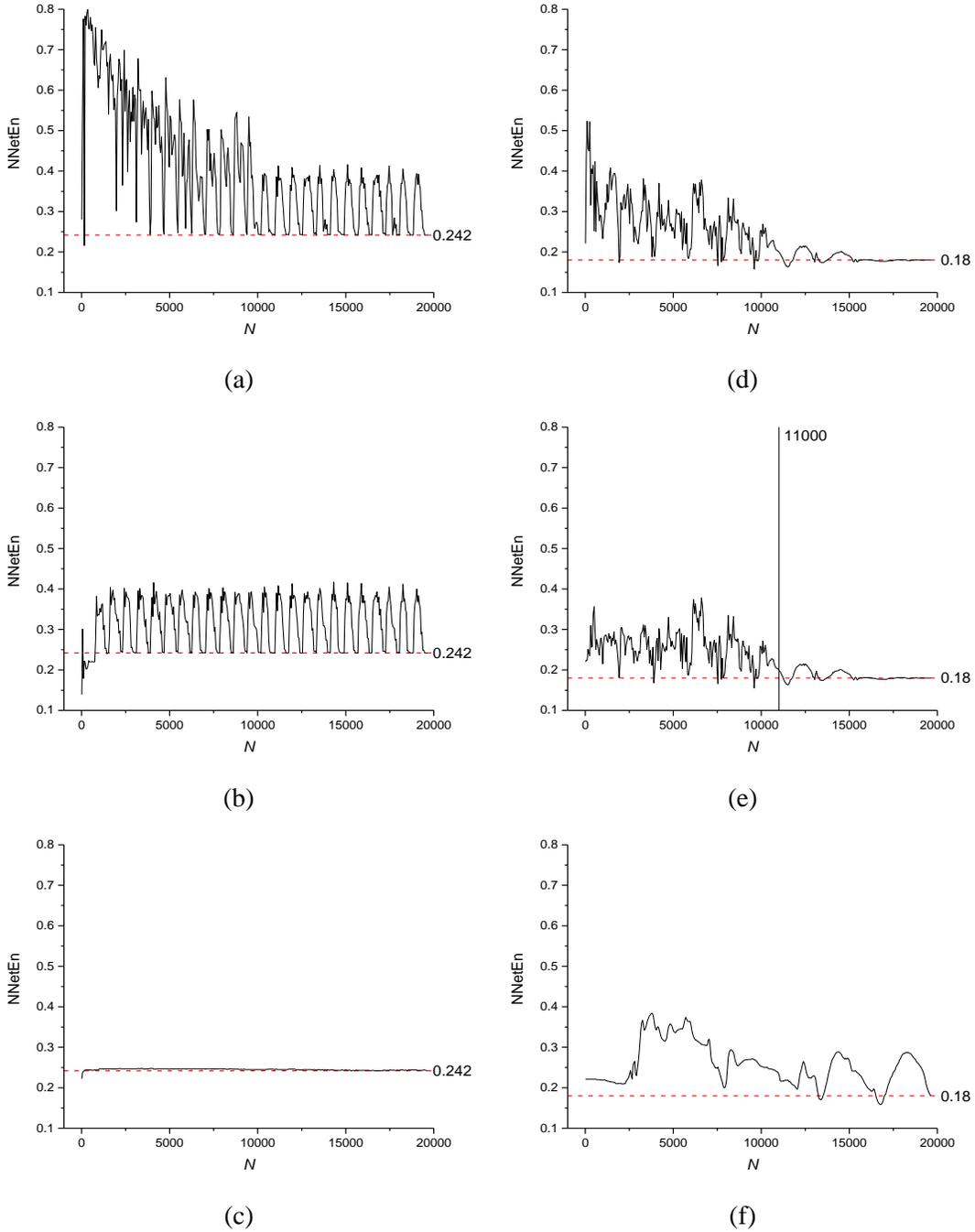

(a)

(d)

(b)

(e)

(c)

(f)

**Figure 8.** The effect of time series length on NNetEn values for periodic time series (Eqn 3) using different matrix filling methods: (a) Method 1 (b) Method 2 (c) Method 3 (d) Method 4 (e) Method 5 (f) Method 6.

Method 5 is most stable for $N \geq 11000$, as the NNetEn values are virtually the same as the reference level. Method 5 is most stable for N ≥ 11000 since NNetEn values are virtually identical to reference levels. Method 1 is suitable for $N < 11000$ because there is less variation and deviation from the reference level. Figure 8 and Figure 9 illustrate the NNetEn measures for a periodic discrete map (Eq.(3)) and a binary discrete map (Eq.(4)) with different numbers of data points in the range $N = 10,..., 19625$.

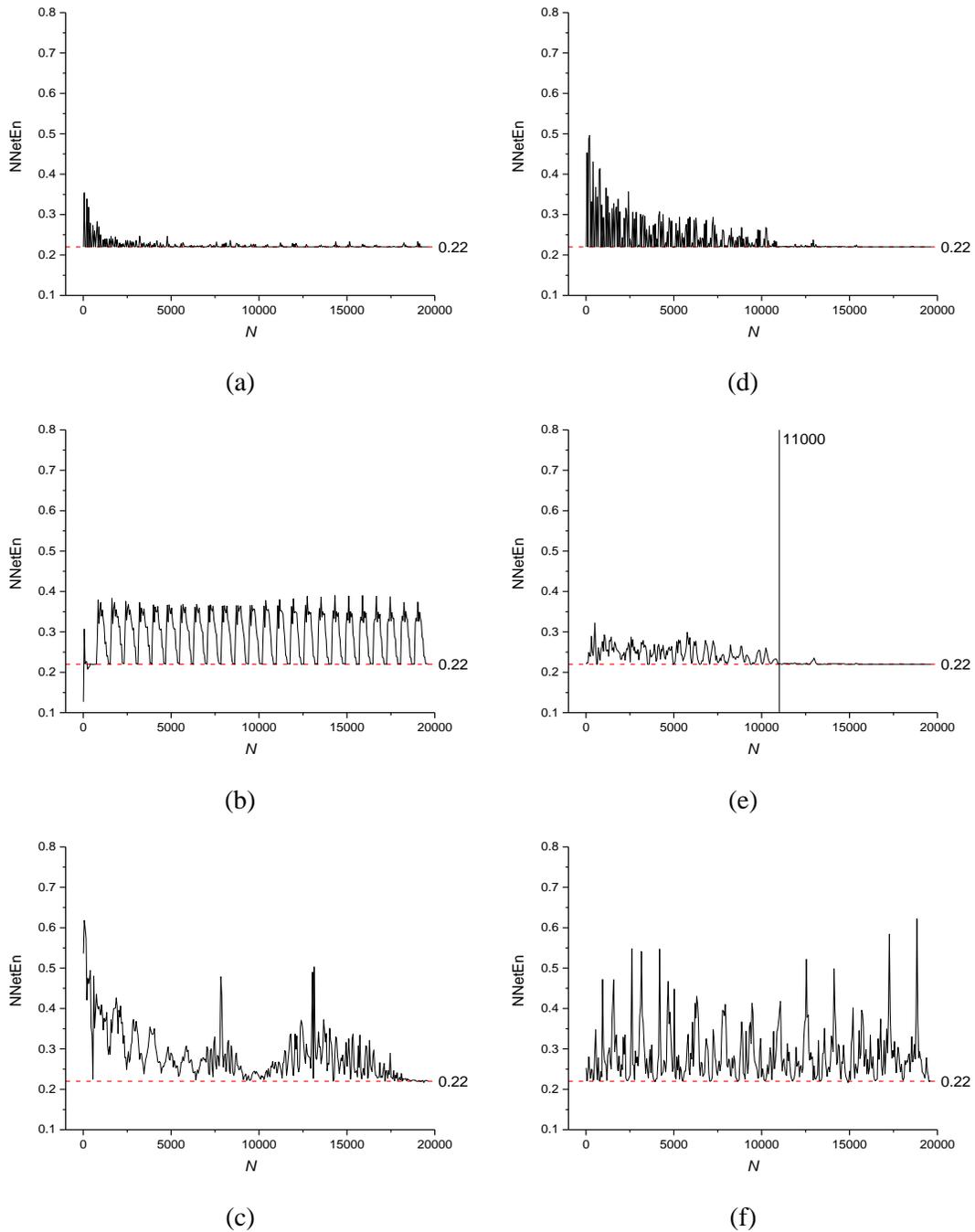

**Figure 9.** The effect of time series length on NNetEn values for binary time series (Eqn.4) using different matrix filling methods: (a) Method 1 (b) Method 2 (c) Method 3 (d) Method 4 (e) Method 5 (f) Method 6.

Method 3 is most suitable for periodic time series with $N > 5$ since it does not fluctuate in entropy values and converges in a small number of cases ($N$). Both Method 3 and Method 6 use the novel stretching approach when intermediate elements have been generated that complement the time series without altering its dynamics. This approach is especially relevant for short series of physical nature.

In Figure 9, method 5 gives the most stable results for binary time series when $N > 11000$, and method 1 is suitable when $N > 5$. Based on Method 2 (row-wise filling with an additional zero element), NNetEn exhibits divergence and periodic behavior for all-time series considered. In a similar comparison, Method 5 (column-wise filling with an additional zero element) shows that NNetEn's periodic behavior decreases with increasing $N$, and eventually converges.

## 3.2. The effects of noise on the NNetEn measure

We considered the noisy signal in the form

$$y_n = x_n + z_n \tag{6}$$

where $x_n$ is the original signal given by Eqn. (3) - Eqn. (5) and $z_n$ is the random noise. The random noise [51] signal is formulated as,

$$z_n = A \cdot (R - 0.5) + B \tag{7}$$

where $R$ is a random number between 0 and 1, $A$ is the amplitude of noise and $B$ is the bias of the noise signal. To investigate the effect of noise on NNetEn measure, we considered the following signal-to-noise ratio [53],

$$SNR = 20 \cdot log_{10}(\frac{A_{signal}}{A_{noise}}) \tag{8}$$

where, $A_{signal}$ and $A_{noise}$ denote the amplitude of the main signal and the amplitude of noise, respectively.

NNetEn considers matrix filling method 5 to be the main reservoir filling technique among the six different types of filling methods. This is because this method performs well in calculating the entropy measure on two different types of time series data among three different types of time series. The values of NNetEn and learning inertia for the original signal without noise and with different offsets B are presented in Figure 10. The dependencies generally take the form of a bell-shaped curve with a maximum of entropy at a certain bias value and a minimum of LI that corresponds to a maximum of NNetEn. In the case of a binary signal, entropy remains unchanged. Increasing the modulus of constant bias results in a decrease in entropy for sinusoidal and logistic signals.

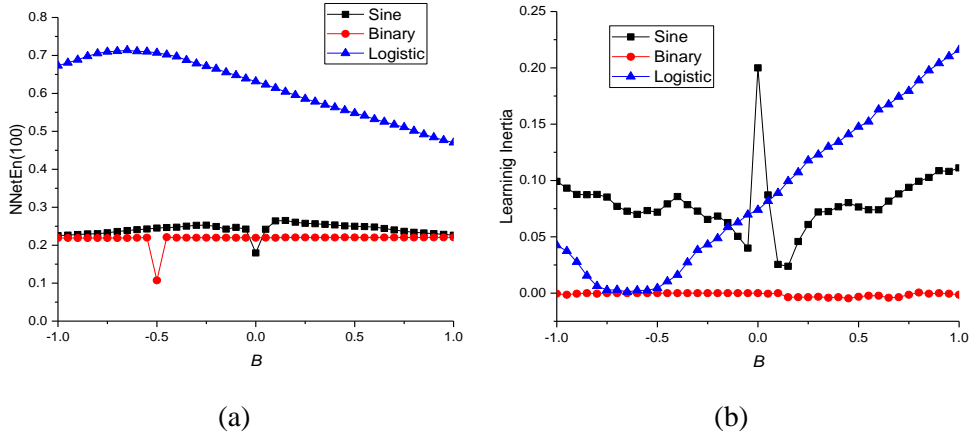

**Figure 10.** (a) The values of NNetEn for the main signals ($A = 0$), (b) The learning inertia values for the main signals.

Different random signals are required for the detection of noise effects on NNetEn. Figure 11 shows the first 100 data points of three different random signals in the form of (Eqn7).

The properties of the random signals are listed in Table 2.

**Table 2.** The statistical properties of random signals.

| Type signal | N total | Mean | Standard deviation | Sum | Minimum | Median | Maximum |
|---|---|---|---|---|---|---|---|
| random_1 | 19625 | 0.0015 | 0.28809 | 29.39545 | -0.49999 | 0.00631 | 0.49999 |
| random_2 | 19625 | 0.00242 | 0.28771 | 47.46514 | -0.49999 | 0.00347 | 0.50000 |
| random_3 | 19625 | -0.00265 | 0.28832 | -52.07462 | -0.49993 | -0.00575 | 0.50000 |

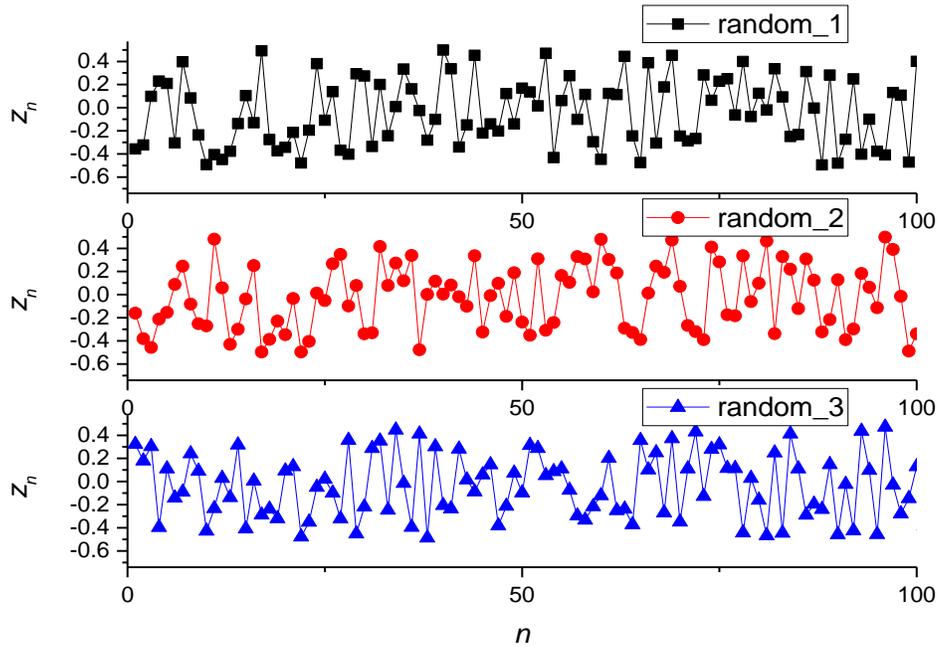

**Figure 11.** The first 100 data points of the random signals of the form (Eqn 7) with $A = 1$ and $B = 0$.

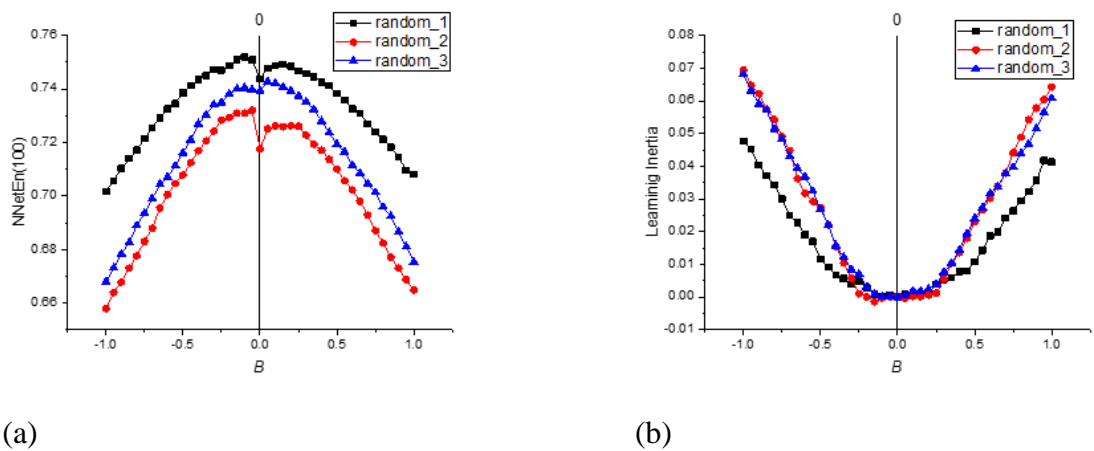

(a)           (b)

**Figure 12.** (a) The values of NNetEn of random signals. (b) The learning inertia of random signals.

Figure 12 shows the NNetEn and learning inertia values for random signals. We found that the NNetEn value for all random signals is a function of parameter B, in the form of an inverted parabola (bell shape) with a maximum at B = 0, while the value of inertia has a minimum value at this point. Due to its small value for all three random signals, the mean value (Table 2) does not play a significant role in the analysis. The complexity of the signals at B = 0 determines the entropy of the signals. As a result of its high entropy and complexity, the signal random_1 is the most complex. Across all three signals, learning inertia is close to zero and increases as |B| increases. It is a universal phenomenon for NNetEn to be dependent on B in a bell-shaped manner. As the bias component B is increased, the role of chaotic components is weakened, and entropy is decreased. In addition, the LI value increases with a decrease in the importance of chaotic components. The LI can be used to assess the degree of chaos in a system in terms of its degree of regularity. Based on our earlier work [49], this may explain the increased LI in the transition region.

### 3.3. Influence of noise amplitude A on the main signal

As discussed in Eqn 7, we examined the effect of noise scale A in NNetEn with B = 0 in this subsection. Figure 13 illustrates how the shape of the sinusoidal signal changes as the noise amplitude increases from $A = 0$ to 1. In the case of $A = 0.05, 0.1$, and 1, the SNR is 26.02 dB, 20 dB, and 0 dB, respectively. An increase in SNR results in a loss of periodicity in the signal as shown in the following figure. Figures 14 through 16 illustrate the effects of amplitude noise on NNetEn for periodic, binary, and chaotic time series. Based on these figures, it can be seen that the value of NNetEn increases as the amount of noise increases, whereas the learning inertia decreases. The SNR for each time series has an upper bound, so the NNetEn measure and *LI* remain unchanged for smaller noise (higher SNR).

The *LI* decreases as the amount of noise increases since the chaotic property of the signal is directly related to the amount of noise. The training intensity increases as the chaotic behavior of the signal increases and, consequently, the *LI* declines. The value of *LI* reaches a pronounced maximum in low-noise areas. Additionally, *LI* decreases significantly as the noise amplitude A increases, especially at low values of *A*, when the signal is approximately periodic.

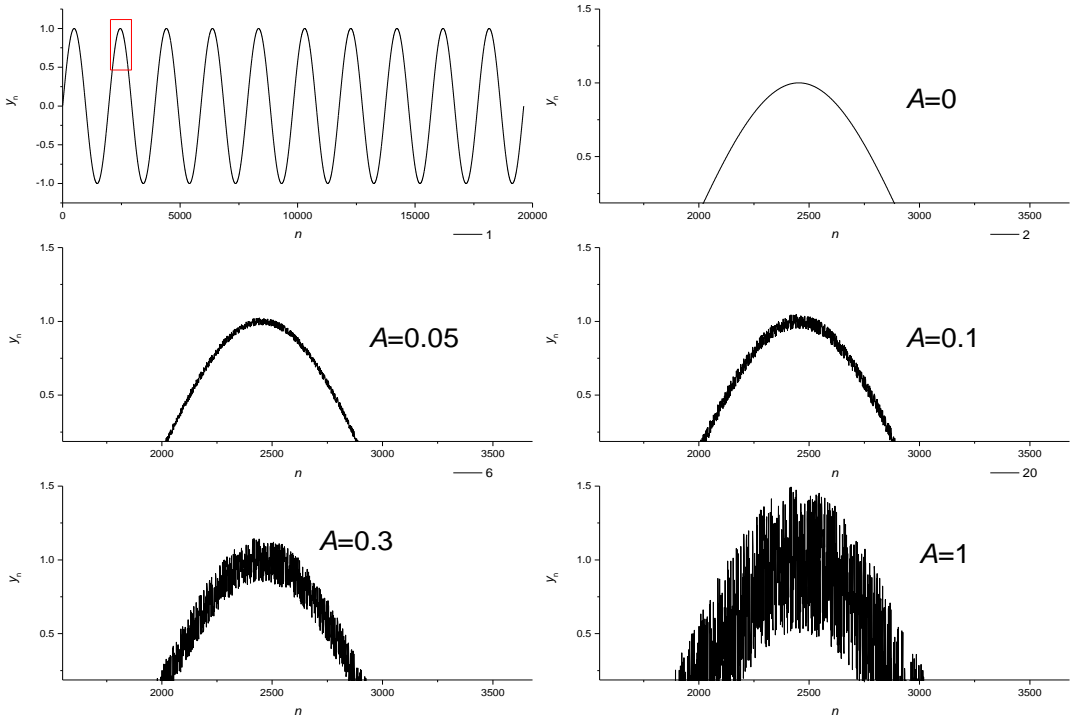

**Figure 13.** The effect of amplitude noise on sine periodic signal.

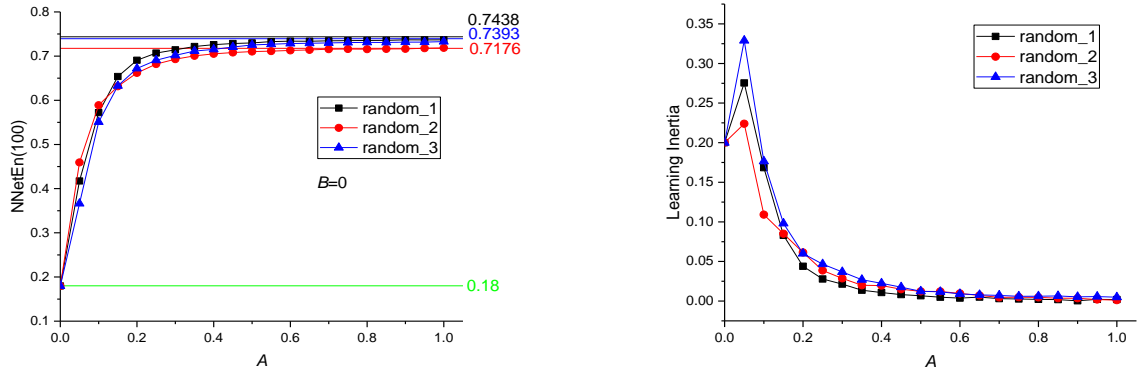

**Figure 14.** The NNetEn (a) and learning inertia (b) values of noisy sine periodic map where the offset is neglected.

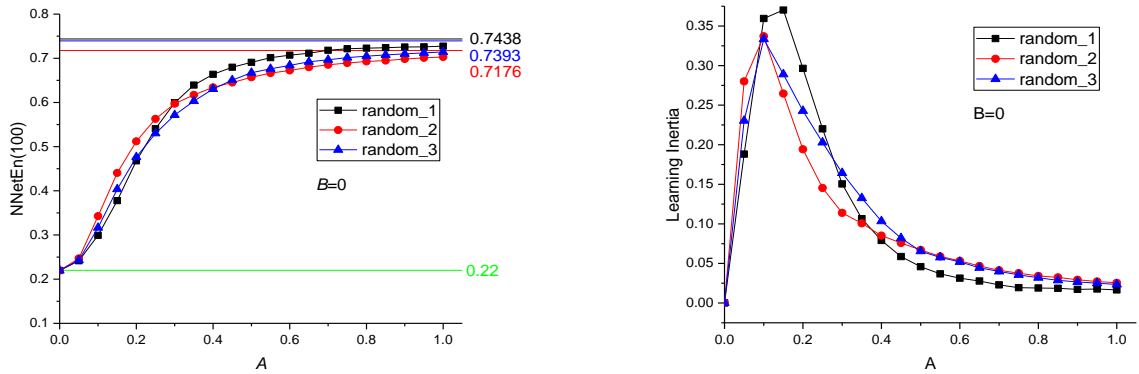

**Figure 15.** The NNetEn (a) and learning inertia (b) values of the noisy binary map where the offset is neglected.

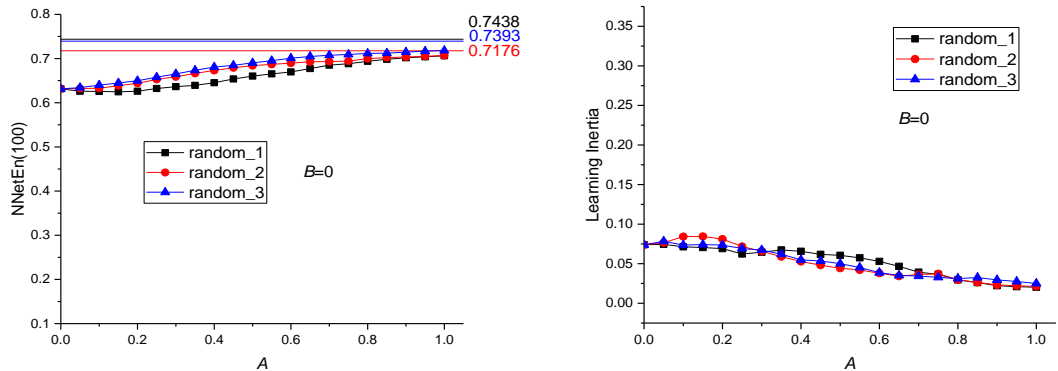

**Figure 16.** The NNetEn (a) and learning inertia (b) values of the noisy logistic map where the offset is neglected.

NNetEn shows the greatest change with increasing *A* when the sine periodic map is considered. In the case of sine, NNetEn increases by 100% when A is raised to 0.05, by 20% for the binary map, and by 5% for the logistic chaotic map when *A* is elevated to 0.05. The logistic chaotic map (Figure 16) exhibits a small variation due to the initial chaotic nature of the signal, and the addition of random noise slightly increases its irregularity. As a result, the error in calculating NNetEn entropy for all types of signals is less than 10% when the SNR exceeds 30 decibels. As a result, NNetEn can now be measured for experimental signals with the noise of different characteristics, white noise, or 1/f noise, without requiring noise filtering.

## 3.4 The effect of offset B

In this subsection, we consider noise in the form of Eqn 7, where *A* is fixed, and B varies between 0 and 1 in increments of 0.05. In order to examine the effect of offset *B*, *A* is assumed to be 0.05 or 0.3. The results are shown in Figure 17 to Figure 19. In all figures, NNetEn decreases with increasing bias (*B*). As a result, the stochastic property of the noisy signal decreases with increasing offset B.

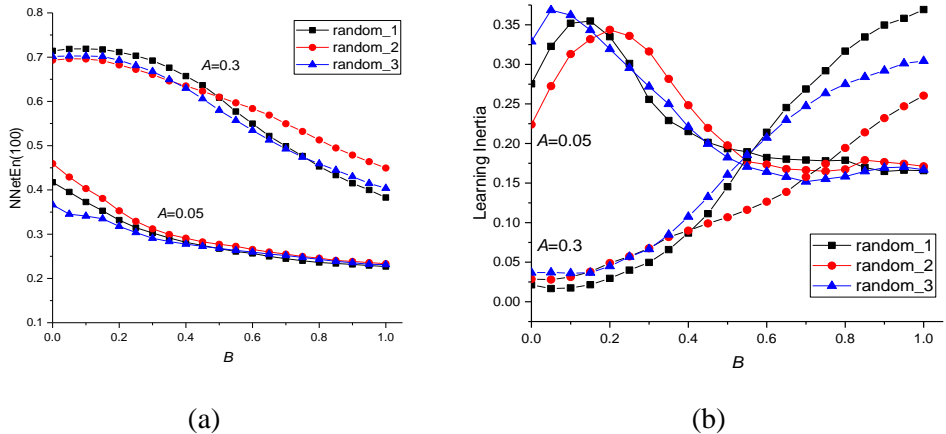

(a)      (b)

**Figure 17.** The NNetEn (a) and learning inertia (b) values of noisy sine periodic map for amplitude noise $A = 0.05$ and 0.3 and different values of offset *B*.

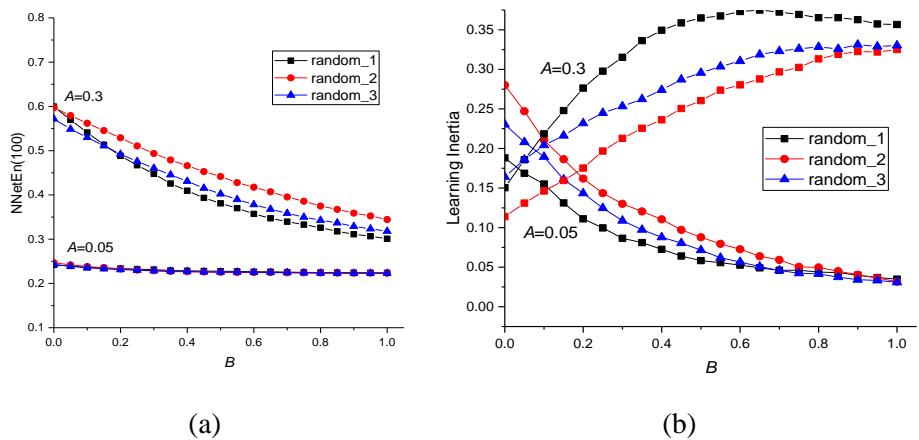

(a)      (b)

**Figure 18.** The NNetEn (a) and learning inertia (b) values of the noisy binary map for amplitude noise $A = 0.05$ and 0.3 and different values of offset *B*.

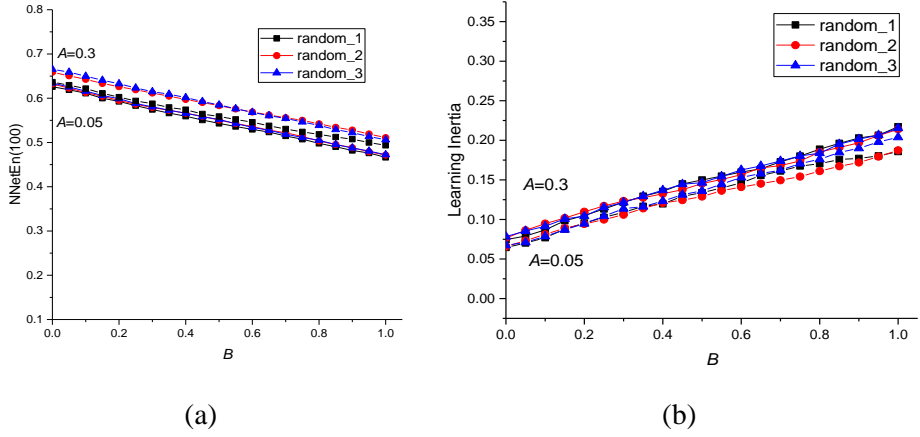

**Figure 19.** The NNetEn (a) and learning inertia (b) values of noisy logistic map for amplitude noise $A = 0.05$ and 0.3 and different values of offset $B$.

There is a considerably higher degree of complexity in the behavior of LI dependent on B than in NNetEn dependent on B. For a sinusoidal signal, there are maxima and minima, and for a logistic chaotic map, an increase in B increases LI. If B is increased by 5%, NNetEn will be affected by 10%. Therefore, B has a smaller effect on NNetEn than A.

**3.5 Analysis of NNetEn with real-time data**

To investigate the performance of NNetEn with real-time data, we examined an EEG signal acquired from a normal control subject using the 14-channel Emotiv EPOC wireless EEG data acquisition device with a sampling rate of 128 samples per second for a duration of 56 seconds (Figure 20a). A detailed description of the data acquisition environment and experimental settings can be found in [46]. To assess the effectiveness of the proposed NNetEn on time-series data analysis, we also considered a real-time example of time-series data collected from an emotion recognition experiment. Rapid technological advancements have made affective computing a hot topic from both a theoretical and practical perspective in recent years. As affective computing research has grown rapidly over the past few years, it has been applied to a variety of fields, including healthcare, robotics, management, marketing, and smart technologies [39, 40]. Different modalities can be used to assess emotion, including facial expressions, gestures, speech signals, and biosignals (EEG, Electromyogram (EMG), Electrocardiogram (ECG), etc.). EEG signals are more robust since they are not affected by fake or mask behaviors and directly correlate with the electrical activity of the brain to detect minute changes in emotion [41, 42]. An EEG signal can be acquired by a low-cost data acquisition device, is more temporally accurate, and offers a high spatial resolution. According to Soroush et al. , different methods have been developed to classify emotions based on EEG signals [39]. Murugappan et al. used extreme machine learning along with recurrent quantification analysis (RQA) for the detection of EEG emotions in Parkinson's disease patients [43]. Li et al introduced the mean differential entropy method for the classification of EEG emotions [44]. In a study conducted by Murugappan et al., a tunable Q-wavelet transform was used in order to improve the accuracy of EEG emotion recognition for patients with Parkinson's disease [42]. A tunable Q-wavelet transform, and the support vector machine method were used by Tuncer et al. to recognize emotion from EEG signals [40]. A study conducted by Li et al investigated the recognition of emotions using EEG in the presence of noise [45]. According to [46], researchers collected the EEG data from 14 normal and healthy subjects for the purpose of detecting six basic emotions: happiness, sadness, fear, anger, disgust, and surprise.

Figure 20b illustrates the spectrum of the signal being analyzed. A large portion of the spectrum covers the range from 0.02 to 30 *Hz*, with zero DC component, while a high-frequency region reaches frequencies of 65 *Hz*.

An IIR Butterworth band-pass filter of 4th order is usually used to filter the raw signal in the range of 0.5 to 49 (Figure 20c). Therefore, it allows for the removal of high-frequency and low-frequency noises due to interference with power lines, movement artifacts, baseline interferences, etc. Typically, high-frequency and low-frequency noises are filtered by low-pass filters (Figure 21a) and high-pass filters (Figure 21b), respectively. Figures 21c and 21d illustrate a typical characteristic of an ideal band-stop and band-pass filter with a lower (Fc1) and upper (Fc2) cut-off frequency. A raw EEG signal has an entropy of NNetEn (100) = 0.637, and after passing through a band-pass filter with a cut-off frequency of 0.5 – 49 *Hz*, it has an entropy of NNetEn(100) = 0.657. The removal of high-frequency and low-frequency noises has not significantly affected the NNetEn (100) entropy value, since the signal under consideration initially exhibits a high entropy in the range of 0.5 - 49 *Hz*.

This section examines the impact of low-pass and high-pass FFT filters on the value of NNetEn (100). Figure 22a and Figure 22b illustrate the effects of low-pass filtering with Fc = 40, 20, 5 *Hz* and high-pass filtering with Fc = 0.1, 0.5, 1 *Hz*. According to the spectral response of the original signal, the cut-off frequencies above have been arbitrarily selected.

The next step is to analyze the effect of low-pass and high-pass FFT filters on the value of NNetEn (100). Figures 22a and 22b illustrate the effects of low-pass filtering with Fc = 40, 20, 5 *Hz* and high-pass filtering with Fc = 0.1, 0.5, 1 *Hz*. Based on the spectral response of the original signal, the above cut-off frequencies were arbitrarily selected.

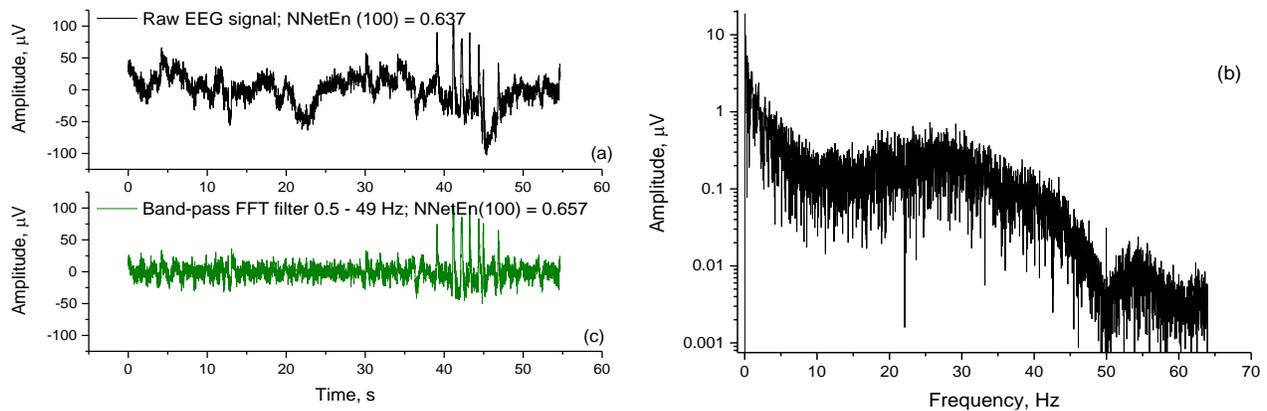

**Figure 20**. An example of a raw EEG signal (a), frequency's spectrum of EEG raw signal (b), signal after band-pass FFT filter 0.5-49 *Hz* (c).

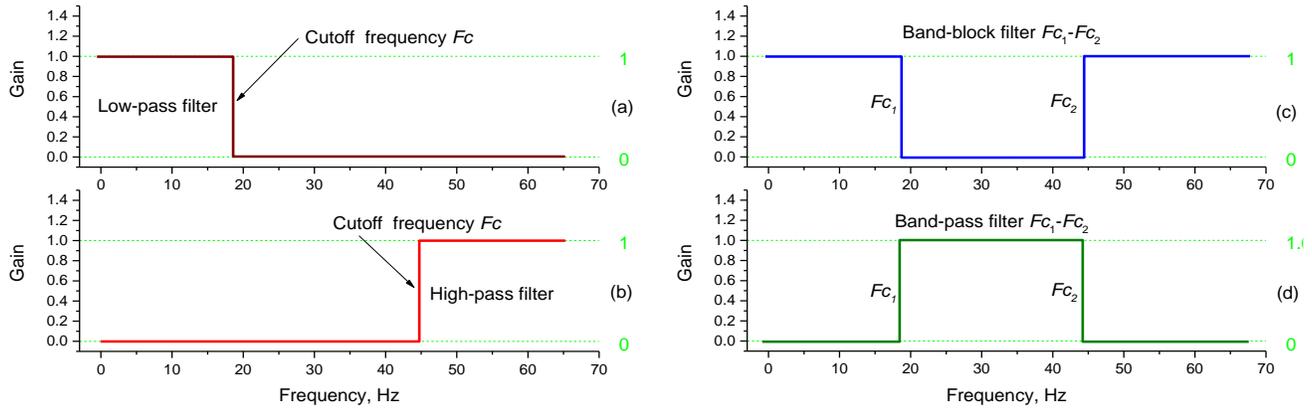

**Figure 21**. Frequency characteristics of low-pass (a), high pass (b), band-stop (c), and band-pass (d) filters

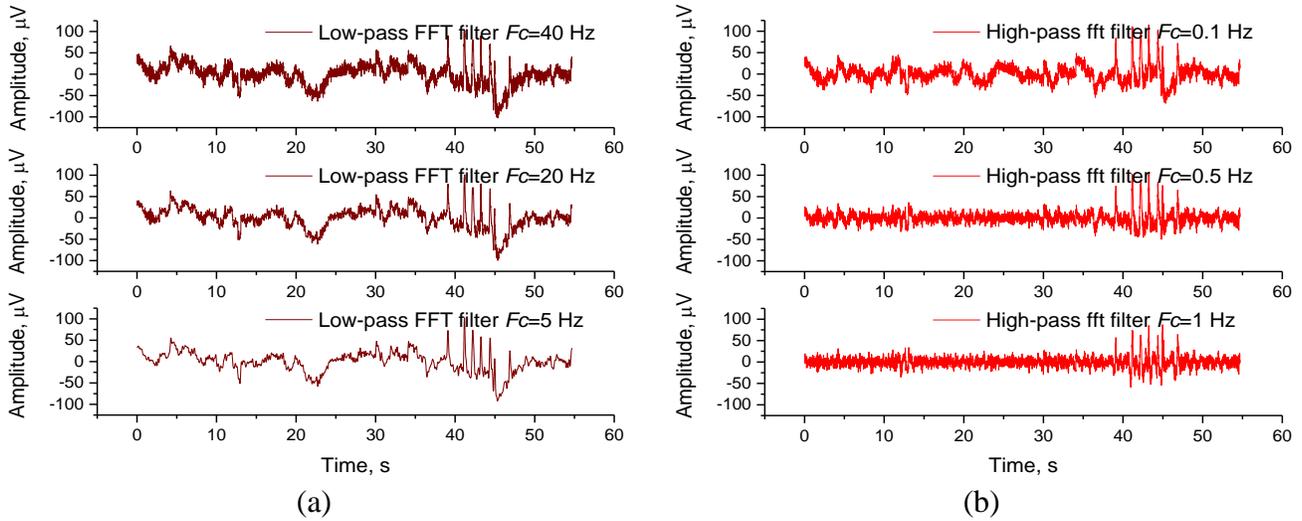

**Figure 22.** Examples of EEG signal filtering when applying (a) Low-pass filter (b) High - pass filter

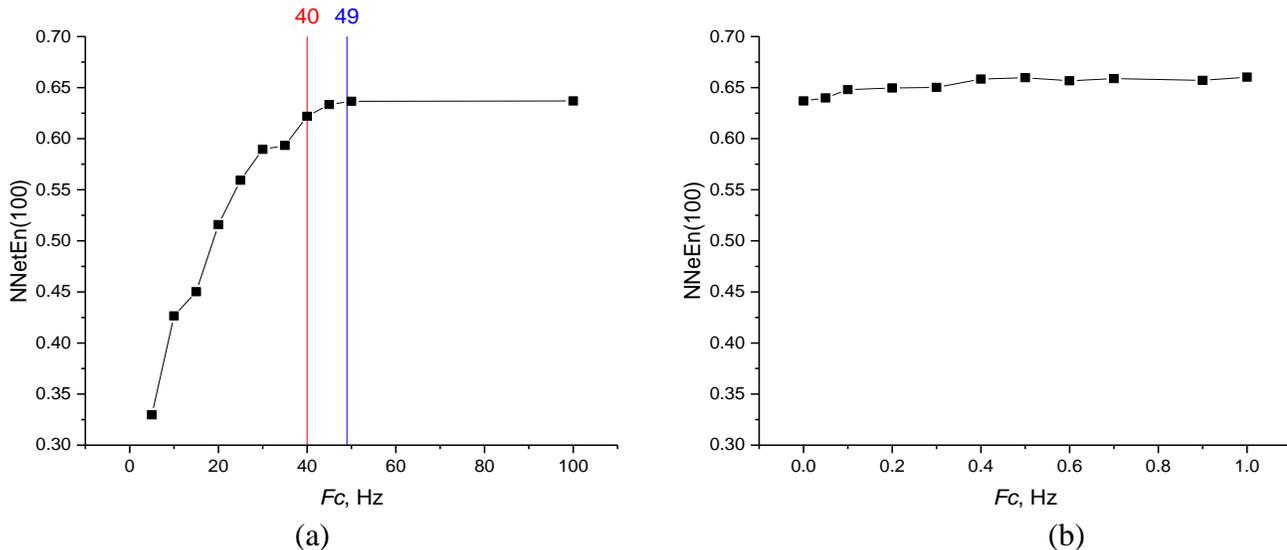

**Figure 23**. NNetEn (100) for EEG signal after applying (a) low-pass filter (b) high-pass filter

The value of NNetEn (100) has been calculated for a low-pass filtered signal with different cut-off frequencies Fc, as shown in Figure 23a. If Fc > 40 *Hz*, NNetEn (100) ~ 0.63 is the same, and practically does not change. Therefore, the presence of high-frequency noise with a frequency of over 40 *Hz* does not affect the value of NNetEn, and the EEG signal entropy can be measured quite accurately without a preliminary filtering process.

Using a low-pass filter with Fc = 40 *Hz* reduces entropy, so the signal becomes smooth, and irregularity is tamed. Figure 22a presents signals that have passed low-pass filters with a cut-off frequency of Fc = 20 *Hz* and Fc = 5 *Hz*, with entropies of NNetEn (100) ~ 0.52 and NNetEn (100) ~ 0.33, respectively. The signal has the least randomness at Fc = 5 *Hz*, due to its smoothed shape compared to raw EEG data. A high-pass filter does not significantly affect NNetEn (100), as shown in figure 23b.

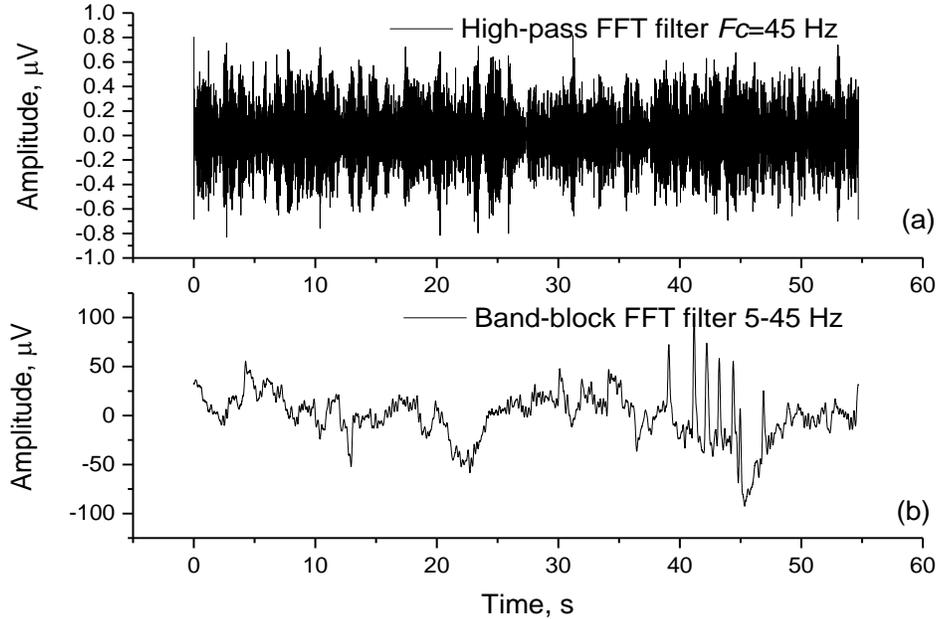

**Figure 24**. Signals: (a) High frequency noise and (b) signal after band-stop filter

Comparing Figures 22b, it is evident that the shape of the signal changes significantly if the low-frequency component is removed. Variations in signal shape are caused by an increase in Fc, which results in an increase in the entropy value (see Figure 23b). It can be explained by the fact that the entropy value increases in high-frequency regions. In Figure 20(a), the sample EEG signal after band-pass filter 0.5 *Hz* and 49 *Hz* had an entropy value NNetEn(100) = 0.657. We examined the robustness of NNetEn of sample EEG signal under a band-block filter. As shown in Figure 23(a), we filtered the EEG signal using an IIR Butterworth low-pass filter with a cut-off frequency of Fc = 5 *Hz* and determined that NNeEn (100) is approximately 0.33. The value of NNetEn (100) decreases to 0.5 when a high-pass filter with a cut-off frequency of Fc = 45 *Hz* is applied (Figure 24a). As shown in Figure 24b, a high-pass and low-pass filter (band-block filter) are combined to produce the resultant signal. A band-block filtered signal (Figure 24b) has an entropy value of NNeEn (100) ~0.345, similar to the entropy of a low pass filtered signal with Fc = 5 *Hz*. Thus, the removal of the main carrier frequencies in the range of 5–45 Hz leads to a decrease in entropy.

As a result, NNetEn measures are robust in the presence of noise. NNetEn allows the analysis of various types of real-time biosignals or time series data using the same method.

## 4. Conclusion

This study proposes a novel NNetEn entropy measure for calculating the entropy values of time-series data of different lengths. NNetEn stores the input time-series data in a reservoir matrix with N = 19625 data points. Six methods are proposed for filling the reservoir of the studied time series of any length $5 \leq N \leq 19625$. Methods 3 and 6 stretch time series by complementing them without affecting their dynamics. Short-time series data are handled more efficiently by Method 3 and Method 5. We have investigated three different time series data types (chaotic, periodic, and binary) with different dynamic properties, assorted signal-to-noise ratios (SNRs), and offset values. Found that NNetEn entropy calculation errors are less than 10% when SNR is greater than 30dB. It is also shown that with an increase in the bias component, the role of the chaotic components is weakened, and entropy decreases. NNetEn calculation methodology has been evaluated using real-time EEG data collected from emotion recognition tasks to assess its robustness under various noise conditions. The experimental results indicate that the NNetEn measures are robust under low-amplitude noise. Thus, NNetEn is an effective measure of entropy when applied to real-world environments with various types of noise. Even though the proposed method retrieves information from the given time series data efficiently, it's computationally expensive (requires more computational memory and time). Our goal in future research is to expand the supported platforms, including Python, and to speed up calculations to reduce their computational complexity. Furthermore, NNetEn computation can also be applied to applications that involve multidimensional time-series data.

## Data Availability

Soft to calculate NNetEn for all 6 methods of matrix filling available in the Researchgate repository, https://www.researchgate.net/publication/366575849_NNetEn_calculator_1004_for_NNetEn_calculation_with_six_matrix_filling_methods

## Funding:
This research did not receive any specific grant from funding agencies in the public, commercial, or not-for-profit sectors.

## Declaration of Competing Interest
The authors declare that they have no known competing financial interests or personal relationships that could have appeared to influence the work reported in this paper.